\documentclass{article}
\usepackage{iclr2021_workshop,times}

\usepackage{amsmath,amsfonts,bm}

\def\eqref#1{equation~\ref{#1}}

\def\1{\bm{1}}

\DeclareMathAlphabet{\mathsfit}{\encodingdefault}{\sfdefault}{m}{sl}
\SetMathAlphabet{\mathsfit}{bold}{\encodingdefault}{\sfdefault}{bx}{n}

\usepackage{graphicx}
\usepackage{hyperref}
\usepackage{url}

\title{Persistent Homology and Graphs Representation Learning}

\author{Mustafa Hajij \\
Department of Mathematics and Computer Science\\
Santa Clara University\\
\texttt{mhajij@scu.edu} \\
\And
Ghaza Zamzmi \\
Department of Computer Science \\
University of South Florida \\
\texttt{ghadh@mail.usf.edu} \\
\AND
Xuanting Cai \\
Facebook Inc \\
\texttt{caixuanting@fb.com}  
}

\iclrfinalcopy 
\begin{document}

\maketitle

\begin{abstract}
This article aims to study the topological invariant properties encoded in node graph representational embeddings by utilizing tools available in persistent homology. Specifically, given a node embedding representation algorithm, we consider the case when these embeddings are real-valued. By viewing these embeddings as scalar functions on a domain of interest, we can utilize the tools available in persistent homology to study the topological information encoded in these representations. Our construction effectively defines a unique persistence-based graph descriptor, on both the graph and node levels, for every node representation algorithm. To demonstrate the effectiveness of the proposed method, we study the topological descriptors induced by DeepWalk, Node2Vec and Diff2Vec. 

\end{abstract}

\section{Introduction}
Graphs are among the most ubiquitous models in computer science with an array of applications in drug discovery (\cite{takigawa2013graph}), biological protein-protein networks (\cite{sun2017sequence}), data representation and organization (\cite{gross2005graph}), recommendation systems (\cite{kutty2014people}), and social networks (\cite{nettleton2013data}). Due to the necessity of operating a typical machine learning pipeline with Euclidean data, there has been a surge of interest in learning graph representation. Graph representation aims to learn a map that embeds the nodes of a graph, or subgraphs of it, to some Euclidean space such that this embedding captures some useful information about the graph. In particular, \textit{node representation learning} (\cite{zhang2018network}) has been mostly inspired by the success of Word2Vec (\cite{mikolov2013efficient}), a
method to embed a corpus of words into the Euclidean space based on their context in the sentences. Similar to Word2Vec, node representation embeddings learn a latent space representation of the nodes of a given graph and are very useful when performing downstream machine learning tasks on graphs. This includes node and graph classification, community detection, link prediction, and graph similarity (\cite{chen2020graph}). Despite the practical popularity of node graph representations, their expressive and discriminative power has not been explored extensively.





This article aims to understand and study the information encoded in node representation embeddings. Specifically, we aim to understand the expressive and discriminative powers as well as the limitations of these embeddings from the perspective of graph classification. Our proposed method utilizes persistent homology (\cite{edelsbrunner2010computational}) to extract the topological invariants from a given latent representation; i.e., for a given node representational embedding $f:V_G\to \mathbb{R}^d $, we are interested in the case when the dimension $d$ of the latent space $\mathbb{R}^d$ is one. In particular, we consider the lower-star filtration obtained from such scalar functions and utilize its induced invariant of $G$, the so called persistence diagrams, obtained via this filtration. The final persistence-based representation that we obtain is a function of the chosen node-embedding algorithm and the input graph.  

Recent years have witnessed an increased interest in the role topology plays in machine learning and data science (\cite{carlsson2009topology}). Topology provides a set of natural tools that allows the formulation of many long-standing problems in these fields. In particular, persistent homology (\cite{edelsbrunner2010computational}) has been successful at finding solutions to a large array of complex problems (\cite{attene2003shape, bajaj1997contour, kweon1994extracting, LeeChungKang2011b, LeeChungKang2011, lum2013extracting, rosen2017using}). These sets of tools have matured and become an area of research known as 
\textit{Topological Data Analysis} (TDA) (\cite{edelsbrunner2010computational,carlsson2009topology}). TDA-based methods have shown excellent performance in several applications including neuroscience~(\cite{ LeeKangChung2012}), bioscience~(\cite{chan2013topology, dewoskin2010applications, hajij2020graph}), in the study of graphs~(\cite{ PetriScolamieroDonato2013, PetriScolamieroDonato2013b}), time-varying data~(\cite{edelsbrunner2004time, hajij2018visual}) among others.

The major contribution of this work is the utilization of the tools available in persistent homology to study the topological information encoded in node embedding representations. For every node representation algorithm, our construction effectively defines a unique graph descriptor. We show the effectiveness of the proposed method by studying the topological descriptors induced by DeepWalk (\cite{perozzi2014deepwalk}), Node2Vec (\cite{grover2016node2vec}) and Diff2Vec (\cite{rozemberczki2018fast}). Our preliminary results are promising and indicate that the information extracted by our method can be effectively utilized for graph classification.

The rest of the paper is organized as follows. Section \ref{background} reviews the necessary tools for our constructions, persistent homology, and graph autoencoders. In Section \ref{main}, we describe the graph topological embeddings obtained using persistent homology and node embeddings. Section \ref{why} is devoted to providing insights on the advantages of our method. Finally, we demonstrate our results in Section \ref{results}.


\section{Setting the Scene}
\label{background}
In this section, we review the necessary tools to build our machinery. Persistent Homology is reviewed in \ref{homology} whereas graph autoencoders are reviewed in \ref{autoencoders}.

\subsection{Persistent Homology}
\label{homology}




Before proceeding, we assume the reader has basic familiarity with simplicial complexes and simplicial homology. We also would like to note that although we only introduce persistent homology for simplicial complexes, this work can be easily extended to more general domains \footnote{AutoEncoders on generalized simplicial complexes are given in (\cite{hajij2020cell}). }. 

Let $K$ be a simplicial complex. We will denote the vertices of $K$ by $V(K)$. Let $S$ be an ordered sequence $\sigma_1,\cdots,\sigma_n$ of all simplices in $K$, such that for any simplex $\sigma \in K$ every face of $\sigma$ appears before $S$. Then $S$ induces a nested sequence of subcomplexes called a \textit{ filtration}:
$\phi=K_0  \subset K_1 \subset ... \subset K_n = K$.
 A $d$-homology class $\alpha \in H_d(K_i)$ is said to be \textit{born} at the time $i$ if it appears for the first time as a homology class in $H_d(K_i)$. A class $\alpha$ \textit{dies} at time $j$ if it is trivial in $H_d(K_j)$ but not trivial in $H_d(K_{j-1})$. The \textit{persistence} of $\alpha$ is defined to be $j-i$. Persistent homology captures the birth and death events in a given filtration and summarizes them in a multi-set structure called the \textit{persistence diagram} $P^d(\phi)$. Specifically, the persistence diagram of the a filtration $\phi$ is a collection of pairs $(i,j)$ in the plane where each $(i,j)$ indicates a $d$-homology class that is created at time $i$ in the filtration $\phi$ and killed entering time $j$. Persistent homology can be defined for \textit{any filtration}. For the purpose of this work, we assume that the input is a piecewise linear function $f:|K|\longrightarrow \mathbb{R}$ defined on the vertices of complex $K$. Further, we assume the function $f$ has different values on different nodes of $K$. Then, any such function induces the \textit{lower-star} filtration as follows. Let $V=\{v_1,\cdots,v_n\}$ be the set of vertices of $K$ sorted in non-decreasing order of their $f$-values, and let $K_i:= \{\sigma \in K | \max_{v \in \sigma}f(v)\leq f(v_i)   \} $. 
The lower-star filtration is defined as: 

\begin{equation}
\label{filter2}
    \mathcal{F}_f(K):  \phi=K_0  \subset K_1 \subset ... \subset K_n = K.
\end{equation} 
The interpretation of the lower-star filtration is rather intuitive; viz., the lower-star filtration reflects the topological information encoded in the scalar function $f$ in the sense that the persistence homology induced by the filtration \ref{filter2} is identical to the persistent homology of the sublevel sets of the function $f$. The lower-star filtration is our tool to extract the topological information encoded in node embeddings. We describe this in the following section.


\subsection{Graph Autoencoders and Node Representation Learning}
\label{autoencoders}

 
Representational learning on graphs has witnessed a surge of interest over the past few years. The main idea is to learn a mapping that embeds the nodes of a graph inside some latent space while preserving the nodes structural information. Although representational learning methods differ in what they learn from data, one may conceptualize various representational learning strategies via \textit{graph autoencoders} (\cite{hamilton2017representation}). We briefly review this abstraction here as it is needed in later sections. 

Let $G(V,E)$ be a graph, possibly weighted. An \textit{encoder} on $G$ is a function of the form:
\begin{equation}
\label{encoder}
enc : V_G\to \mathbb{R}^d.
\end{equation}

This encoder associates to every node $v$ in $G$ a feature vector $\textbf{z}_v \in \mathbb{R}^d$ that encodes the structure of $v$ and its relationship to other nodes in $G$. A \textit{decoder} is a function of the form: $
    dec : \mathbb{R}^d \times \mathbb{R}^d\to \mathbb{R}^+.
$
The decoder associates to every pair of node embeddings a similarity score that quantifies some notion of relationship between these nodes. The pair $(enc,dec)$ on $G$ is called a \textit{graph AutoEncoder} on $G$. The functions $enc$ and $dec$ are typically trainable functions that are optimized using user-defined similarity measure and loss function. The abstraction above encompasses most popular method for node representation learning including Graph Factorization (\cite{angles2008survey}), Node2Vec (\cite{grover2016node2vec}), and DeepWalk (\cite{perozzi2014deepwalk}). In our context, we are interested in the case when the latent space in (\ref{encoder}) is $\mathbb{R}$. Specifically, we are interested in the case when the encoder $enc:V_G\to \mathbb{R}$ is a scalar function on $G$. By utilizing the lower-star filtration on an encoder of interest, we obtain a topological descriptor on the graph $G$ that reflects the topological information encoded in this learned representation.

\section{Obtaining topological embeddings using Persistent Homology and Node Representation Learning
 }
 \label{main}
 
 Given a learned node representation $enc: V_G\to \mathbb{R} $ as described in the previous section, one may extract the topological information encoded by computing the persistence diagrams of its lower-star filtration. The persistence diagram can then be converted to a feature vector inside some feature space to perform a downstream machine learning task of interest. This process of converting the persistence-based representation to a vector is called vectorization of the persistence diagram. Many vectorization schemes have been suggested recently. Examples of these schemes include betti curve (\cite{umeda2017time}), the persistence landscape (\cite{bubenik2015statistical}), the persistence image (\cite{adams2017persistence}) and many other (\cite{chen2015statistical,berry2020functional,kusano2016persistence}). In our context, the persistence diagram of a learned representation embedding can be converted to a vector using any of the vectorization methods referenced above.   
\subsection{Persistence-based Node Embeddings}
The above procedure can be altered to obtain a persistence-based node embedding that induces an  encoder of the form \ref{encoder}. To this end, we need to introduce the \textit{ego-network} of a vertex. Let $w$ be given vertex $G$. The \textit{$k$-hop ego network} of $w$, is a subgraph $G^k_w(\mathcal{N}^k(w),E^k_w)$ of $G$. The vertex set $\mathcal{N}^k(w)$ consists of the node $w$ and all other nodes in $V_G$ that can be reached in $k$ steps from $w$. The edges $E^k_w$ consists of  every edge of $G$ whose endpoints are in $\mathcal{N}^k(w)$. Given a node $w \in V_G $ and an encoder $enc$ as described in \ref{encoder}, one may define the following restriction:
$
enc_{w}: \mathcal{N}^k(w)\to \mathbb{R}
$, and then compute the lower-star filtration on this restriction to obtain a node-level topological representation of the node $w$.

One might wonder why we need to create the node-level persistence-based representation for a node $w$ and what are the advantages of having this representation over the ''raw'' representation already contained in the encoder? We provide a justification for this point next.




\section{Why Using Persistent Homology To obtain Descriptors from Node embeddings ?}
\label{why}

Node representation embeddings naturally induce graph descriptors by directly using the node embedding readily available from the representation algorithm. For instance, one may simply add all node embeddings of a given graph to obtain a graph descriptor. This begs for the question, why do we need to utilize persistent homology to obtain a graph descriptor from the node embeddings?   

Recently, there has been multiple efforts to provide a partial answer for the above question. For example, several studies \cite{nielson2015topological,joshi2019survey} reported that the topological features extracted by the persistence diagram can 1) track different information from the original encodings obtained from the raw data \cite{dey2017improved,hajij2021tda} and 2) determine which predictors are more related to the outcome. In addition, the topological analysis of a graph allows to extract features that are invariant to the spatial transformation and more robust to noise \cite{zheng2015application,bae2017beyond}. For instance, DeepWalk and Node2Vec are stochastic node embeddings by design and different runs of the algorithm on the same graph yield different outputs. Our results in Section \ref{results} indicates that while this is the case, the topological information extracted from the persistence diagram yielded a feature vector that is robust against such stochasticity. See Section \ref{results} for more details. 

In addition, the fact that persistence homology tracks different information from the original deep learning-based representations (\cite{dey2017improved,hajij2021tda}) suggests that combining the encodings obtained from a persistent homology and other invariants obtained directly from node embedding should increase the quality of the learning task at hand. Further, persistent homology provides a clear, concise and rigorous method to quantify the level of expressiveness of the embeddings as we shall see in Section \ref{results}.  
 

\section{Preliminary Results}
\label{results}
\begin{figure}[h!]
\begin{center}
\includegraphics[width=1\linewidth]{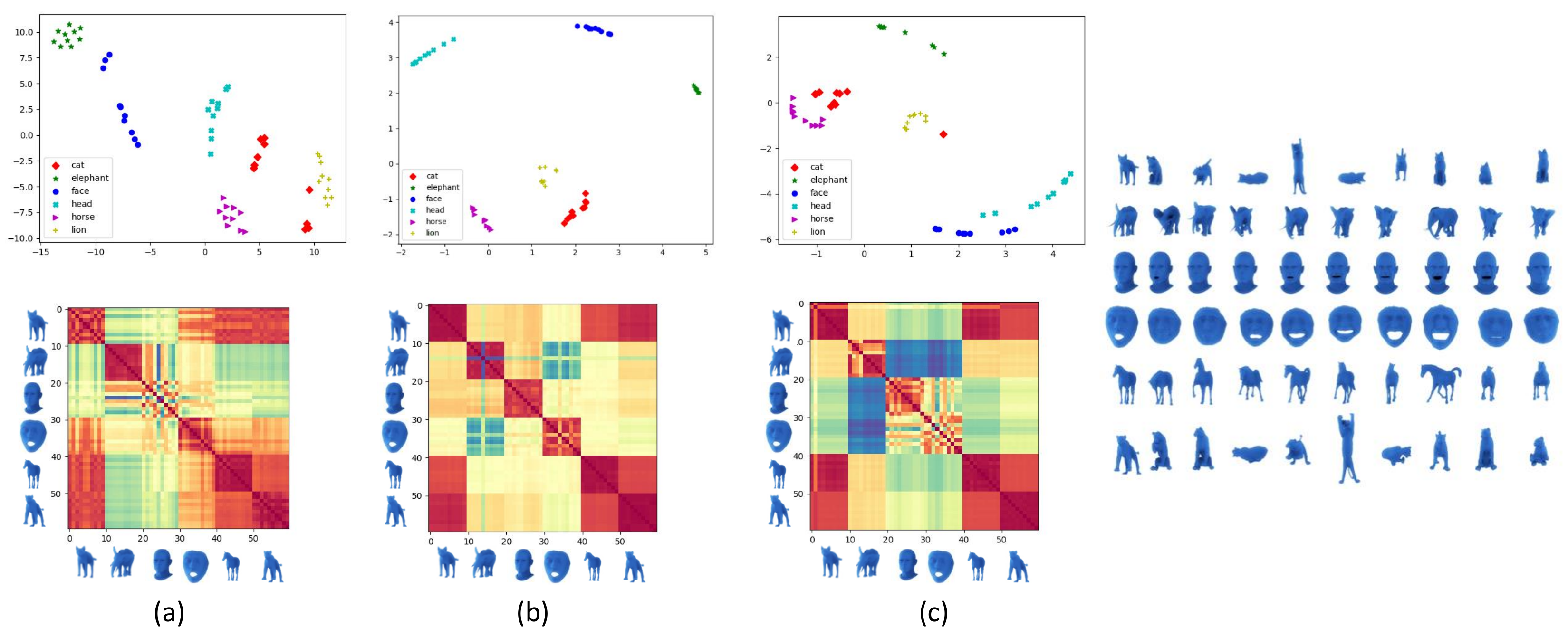}
\end{center}
\caption{ Figure (a) shows the 2d-TSNE projection on the top and the Wasserstein distance matrix on the bottom for Node2Vec, whereas figure (b) and (c) show the results for DeepWalk and Diff2Vec respectively. Note that each descriptor obtained this way clearly provides us with distinct clustering on the level of the metric space. Furthermore, meshes with similar structures, such as horses and lions have clusters that are close to each other. }
\label{Figure}
\end{figure}

To evaluate the proposed method, we used a mesh dataset (\cite{sumner2004deformation}) that consists of 60 meshes divided into
6 categories: cat, elephant, face, head, horse, and lion (see Figure \ref{Figure}).  Each category contains ten triangulated meshes\footnote{Each mesh in the dataset is converted to a weighted graph where the weight on an edge is simply the Euclidean distance between its nodes.}. We compared the proposed method with three node representation embeddings: DeepWalk \cite{perozzi2014deepwalk}, Node2Vec \cite{grover2016node2vec} and Diff2Vec \cite{rozemberczki2018fast}.

For each mesh in the dataset, we compute the node representation vectors and then compute their respective $0- $persistence diagrams. Given two persistence diagrams obtained using the same node representation embedding, we measure the distance between them using the Wasserstein distance. Namely, given two persistence diagrams $X$ and $Y$ obtained from the same node representation embedding, we compute the Wasserstein distance between $X$ and $Y$: 
$
W_q(X,Y) = \displaystyle \left[ \inf_{\eta:X \rightarrow Y}  \Sigma_{x\in X} \left\lVert x-\eta(x) \right\rVert^q_\infty \right]^{1/q}, 
$ 
where $\eta$ is a bijection between points in the diagrams. In our experiment, we set $q=2$. The results are reported in Figure \ref{Figure}. This procedure gives us a Wasserstein distance matrix on the persistence diagrams induced by the lower-star filtration of the node-representation embedding method. Finally, we visualized the resulting discrete metric space using a 2d t-SNE projection (\cite{vanDerMaaten2008}) as shown in Figure \ref{Figure}. These results are promising and suggest that the information extracted in this method can be effectively utilized for a downstream machine learning pipeline on graphs. Moreover, observe that this study demonstrates that the expressive power of the obtained representation varies across the three chosen node embeddings.


 

\label{headings}
\bibliography{iclr2021_workshop}
\bibliographystyle{iclr2021_workshop}


\end{document}